# A syntax-based part-of-speech analyser

Atro Voutilainen
Research Unit for Multilingual Language Technology
P.O. Box 4
FIN-00014 University of Helsinki
Finland
Atro.Voutilainen@Helsinki.FI

### Abstract

There are two main methodologies for constructing the knowledge base of a natural language analyser: the linguistic and the data-driven. Recent state-of-the-art part-of-speech taggers are based on the data-driven approach. Because of the known feasibility of the linguistic rule-based approach at related levels of description, the success of the data-driven approach in part-of-speech analysis may appear surprising. In this paper[1], a case is made for the syntactic nature of part-of-speech tagging. A new tagger of English that uses only linguistic distributional rules is outlined and empirically evaluated. Tested against a benchmark corpus of 38,000 words of previously unseen text, this syntax-based system reaches an accuracy of above 99%. Compared to the 95-97% accuracy of its best competitors, this result suggests the feasibility of the linguistic approach also in part-of-speech analysis.

## 1 Introduction

Part-of-speech analysis usually consists of (i) introduction of ambiguity (lexical analysis) and (ii) disambiguation (elimination of illegitimate alternatives). While introducing ambiguity is regarded as relatively straightforward, disambiguation is known to be a difficult and controversial problem. There are two main methodologies: the linguistic and the data-driven.

- In the linguistic approach, the generalisations are based on the linguist's (potentially corpus-based) abstractions about the paradigms and syntagms of the language. Distributional generalisations are manually coded as a grammar, a system of constraint rules used for discarding contextually illegitimate analyses. The linguistic approach is labour-intensive: skill and effort is needed for writing an exhaustive grammar.

---

[1] This paper is published in *Proceedings of the Seventh Conference of the European Chapter of the Association for Computational Linguistics*, Dublin, 1995.

- In the data-driven approach, frequency-based information is automatically derived from corpora. The learning corpus can consist of plain text, but the best results seem achievable with annotated corpora (Merialdo 1994; Elworthy 1994). This corpus-based information typically concerns sequences of 1-3 tags or words (with some well-known exceptions, e.g. Cutting *et al.* 1992). Corpus-based information can be represented e.g. as neural networks (Eineborg and Gambäck 1994; Schmid 1994), local rules (Brill 1992), or collocational matrices (Garside 1987). In the data-driven approach, no human effort is needed for rule-writing. However, considerable effort may be needed for determining a workable tag set (cf. Cutting 1994) and annotating the training corpus.

At the first flush, the linguistic approach may seem an obvious choice. A part-of-speech tagger's task is often illustrated with a noun–verb ambiguous word directly preceded by an unambiguous determiner (e.g. *table* in *the table*). This ambiguity can reliably be resolved with a simple and obvious grammar rule that disallows verbs after determiners.

Indeed, few contest the fact that reliable linguistic rules can be written for resolving some part-of-speech ambiguities. The main problem with this approach seems to be that resolving part-of-speech ambiguities on a large scale, without introducing a considerable error margin, is very difficult at best. At least, no rule-based system with a convincing accuracy has been reported so far.[2]

As a rule, data-driven systems rely on statistical generalisations about short sequences of words or tags. Though these systems do not usually employ information about long-distance phenomena or the linguist's abstraction capabilities (e.g. knowledge about what is relevant in the context), they tend to reach a 95-97% accuracy in the analysis of several languages, in particular English (Marshall 1983; Black *et al.* 1992; Church 1988; Cutting *et al.* 1992; de Marcken 1990; DeRose 1988; Hindle 1989; Merialdo 1994; Weischedel *et al.* 1993; Brill 1992; Samuelsson 1994; Eineborg and Gambäck 1994, etc.). Interestingly, no significant improvement beyond the 97% "barrier" by means of purely data-driven systems has been reported so far.

In terms of the accuracy of known systems, the data-driven approach seems then to provide the best model of part-of-speech distribution. This should appear a little curious because very competitive results have been achieved using the linguistic approach at related levels of description. With respect to computational morphology, witness for instance the success of the Two-Level paradigm introduced by Koskenniemi (1983): extensive morphological descriptions have been made of more than 15 typologically different languages (Kimmo Koskenniemi, personal communication). With regard to computational syntax, see for instance (Güngördü and Oflazer 1994; Hindle 1983; Jensen, Heidorn and Richardson (eds.) 1993; McCord 1990; Sleator and Temperley 1991; Alshawi (ed.) 1992; Strzalkowski 1992). The present success of the statistical approach in part-of-speech analysis seems then to form an exception to the general feasibility of the rule-based linguistic approach. Is the level of parts of speech somehow different, perhaps less rule-governed, than related levels?[3]

---

[2]There is one potential exception: the rule-based morphological disambiguator used in the English Constraint Grammar Parser ENGCG (Voutilainen, Heikkilä and Anttila 1992). Its recall is very high (99.7% of all words receive the correct morphological analysis), but this system leaves 3-7% of all words ambiguous, trading precision for recall.

We do not need to assume this idiosyncratic status entirely. The rest of this paper argues that also parts of speech can be viewed as a rule-governed phenomenon, possible to model using the linguistic approach. However, it will also be argued that though the distribution of parts of speech can to some extent be described with rules specific to this level of representation, a more natural account could be given using rules overtly about the form and function of essentially *syntactic* categories. A syntactic grammar appears to predict the distribution of parts of speech as a "side effect". In this sense parts of speech seem to differ from morphology and syntax: their status as an independent level of linguistic description appears doubtful.

Before proceeding further with the main argument, consider three very recent hybrids – systems that employ linguistic rules for resolving some of the ambiguities before using automatically generated corpus-based information: collocation matrices (Leech, Garside and Bryant 1994), Hidden Markov Models (Tapanainen and Voutilainen 1994), or syntactic patterns (Tapanainen and Järvinen 1994). What is interesting in these hybrids is that they, unlike purely data-driven taggers, seem capable of exceeding the 97% barrier: all three report an accuracy of about 98.5%.[4] The success of these hybrids could be regarded as evidence for the syntactic aspects of parts of speech.

However, the above hybrids still contain a data-driven component, i.e. it remains an open question whether a tagger entirely based on the linguistic approach can compare with a data-driven system. Next, a new system with the following properties is outlined and evaluated:

- The tagger uses only linguistic distributional rules.

- Tested against a 38,000-word corpus of previously unseen text, the tagger reaches a better accuracy than previous systems (over 99%).

---

[3]For related discussion, cf. Sampson (1987) and Church (1992).

[4]However, CLAWS4 (Leech, Garside and Bryant 1994) leaves some ambiguities unresolved; it uses portmanteau tags for representing them.

- At the level of linguistic abstraction, the grammar rules are essentially syntactic. Ideally, part-of-speech disambiguation should fall out as a "side effect" of syntactic analysis.

Section 2 outlines a rule-based system consisting of the ENGCG tagger followed by a finite-state syntactic parser (Voutilainen and Tapanainen 1993; Voutilainen 1994) that resolves remaining part-of-speech ambiguities as a side effect. In Section 3, this rule-based system is tested against a 38,000-word corpus of previously unseen text. Currently tagger evaluation is only becoming standardised; the evaluation method is accordingly reported in detail.

# 2 System description

The tagger consists of the following sequential components:

- Tokeniser
- ENGCG morphological analyser
  - Lexicon
  - Morphological heuristics
- ENGCG morphological disambiguator
- Lookup of alternative syntactic tags
- Finite state syntactic disambiguator

## 2.1 Morphological analysis

The tokeniser is a rule-based system for identifying words, punctuation marks, document markers, and fixed syntagms (multiword prepositions, certain compounds etc.).

The morphological description consists of two rule components: (i) the lexicon and (ii) heuristic rules for analysing unrecognised words.

The English Koskenniemi-style lexicon contains over 80,000 lexical entries, each of which represents all inflected and some derived surface forms. The lexicon employs 139 tags mainly for part of speech, inflection and derivation; for instance:

```
"<that>"
   "that" <**CLB> CS
   "that" DET CENTRAL DEM SG
   "that" ADV
   "that" PRON DEM SG
   "that" <Rel> PRON SG/PL
```

The morphological analyser produces about 180 different tag combinations. To contrast the ENGCG morphological description with the well-known Brown Corpus tags: ENGCG is more distinctive in that a part-of-speech distinction is spelled out in the description of (i) determiner–pronoun, (ii) preposition–conjunction, (iii) determiner–adverb–pronoun, and (iv) subjunctive–imperative–infinitive–present tense homographs. On the other hand, ENGCG does not spell out part-of-speech ambiguity in the description of (i) -*ing* and nonfinite -*ed* forms, (ii) noun–adjective homographs with similar core meanings, or (iii) abbreviation–proper noun–common noun homographs.

"Morphological heuristics" is a rule-based module for the analysis of those 1–5% of input words not represented in the lexicon. This module employs ordered hand-grafted rules that base their analyses on word shape. If none of the pattern rules apply, a nominal reading is assigned as a default.

## 2.2   ENGCG disambiguator

A Constraint Grammar can be viewed as a collection[5] of pattern–action rules, no more than one for each ambiguity-forming tag. Each rule specifies one or more context patterns, or "constraints", where the tag is illegitimate. If any of these context patterns are satisfied during disambiguation, the tag is deleted; otherwise it is left intact. The context patterns can be local or global, and they can refer to ambiguous or unambiguous analyses. During disambiguation, the context can become less ambiguous. To help a pattern defining an unambiguous context match, several passes are made over the sentence during disambiguation.

The current English grammar contains 1,185 linguistic constraints on the linear order of morphological tags. Of these, 844 specify a context that extends beyond the neighboring word; in this limited sense, 71% of the constraints are global. Interestingly, the constraints are partial and often negative paraphrases of 23 general, essentially syntactic generalisations about the form of the noun phrase, the prepositional phrase, the finite verb chain etc. (Voutilainen 1994).

The grammar avoids risky predictions, therefore 3-7% of all words remain ambiguous (an average 1.04-1.08 alternative analyses per output word). On the other hand, at least 99.7% of all words retain the correct morphological analysis. Note in passing that the ratio 1.04-1.08/99.7% compares very favourably with other systems; c.f. 3.0/99.3% by POST (Weischedel *et al.* 1993) and 1.04/97.6% or 1.09/98.6% by de Marcken (1990).

There is an additional collection of 200 optionally applicable heuristic constraints that are based on simplified linguistic generalisations. They resolve

---

[5]Actually, it is possible to define additional heuristic rule collections that can optionally be applied after the more reliable ones for resolving remaining ambiguities.

about half of the remaining ambiguities, increasing the overall error rate to about 0.5%.

Most of even the remaining ambiguities are structurally resolvable. ENGCG leaves them pending mainly because it is prohibitively difficult to express certain kinds of structural generalisation using the available rule formalism and grammatical representation.

## 2.3 Syntactic analysis

### 2.3.1 Finite-State Intersection Grammar

Syntactic analysis is carried out in another reductionistic parsing framework known as Finite-State Intersection Grammar (Koskenniemi 1990; Koskenniemi, Tapanainen and Voutilainen 1992; Tapanainen 1992; Voutilainen and Tapanainen 1993; Voutilainen 1994). A short introduction:

- Also here syntactic analysis means resolution of structural ambiguities. Morphological, syntactic and clause boundary descriptors are introduced as ambiguities with simple mappings; these ambiguities are then resolved in parallel.

- The formalism does not distinguish between various types of ambiguity; nor are ambiguity class specific rule sets needed. A single rule often resolves all types of ambiguity, though superficially it may look e.g. like a rule about syntactic functions.

- The grammarian can define constants and predicates using regular expressions. For instance, the constants "." and ".." accept any features within a morphological reading and a finite clause (that may even contain centre-embedded clauses), respectively. Constants and predicates can be used in rules, e.g. implication rules that are of the form

  ```
  X =>
  LC1 _ RC1,
  LC2 _ RC2,
  ...
  LCn _ RCn;
  ```

  Here *X, LC1, RC1, LC2* etc. are regular expressions. The rule reads: "X *is legitimate only if it occurs in context* LC1 _ RC1 *or in context* LC2 _ RC2 *... or in context* LCn _ RCn".

- Also the ambiguous sentences are represented as regular expressions.

- Before parsing, rules and sentences are compiled into deterministic finite-state automata.

- Parsing means intersecting the (ambiguous) sentence automaton with each rule automaton. Those sentence readings accepted by all rule automata are proposed as parses.

- In addition, heuristic rules can be used for ranking alternative analyses accepted by the strict rules.

### 2.3.2 Grammatical representation

The grammatical representation used in the Finite State framework is an extension of the ENGCG syntax. Surface-syntactic grammatical relations are encoded with dependency-oriented functional tags. Functional representation of phrases and clauses has been introduced to facilitate expressing syntactic generalisations. The representation is introduced in (Voutilainen and Tapanainen 1993; Voutilainen 1994); here, only the main characteristics are given:

- Each word boundary is explicitly represented as one of five alternatives:

  - the sentence boundary "@@"
  - the boundary separating juxtaposed finite clauses "@/"
  - centre-embedded (sequences of) finite clauses are flanked with "@<" and "@>"
  - the plain word boundary "@"

- Each word is furnished with a tag indicating a surface-syntactic function (subject, premodifier, auxiliary, main verb, adverbial, etc.). All main verbs are furnished with two syntactic tags, one indicating its main verb status, the other indicating the function of the clause.

- An explicit difference is made between finite and nonfinite clauses. Members in nonfinite clauses are indicated with lower case tags; the rest with upper case.

- In addition to syntactic tags, also morphological, e.g. part-of-speech tags are provided for each word. Let us illustrate with a simplified example.

```
                                @@
Mary        N       @SUBJ        @
told        V       @MV    MC@    @
the         DET     @>N          @
fat         A       @>N          @
butcher's   N       @>N          @
```

```
wife        N      @IOBJ           @
and         CC     @CC             @
daughters   N      @IOBJ           @/
that        CS     @CS             @
she         PRON   @SUBJ           @
remembers   V      @MV      OBJ@   @
seeing      V      @mv      OBJ@   @
a           DET    @>N             @
dream       N      @obj            @
last        DET    @>N             @
night       N      @ADVL           @
@fullstop                          @@
```

Here *Mary* is a subject in a finite clause (hence the upper case); *told* is a main verb in a main clause; *the*, *fat* and *butcher's* are premodifiers; *wife* and *daughters* are indirect objects; *that* is a subordinating conjunction; *remembers* is a main verb in a finite clause that serves the Object role in a finite clause (the regent being *told*); *seeing* is a main verb in a nonfinite clause (hence the lower case) that also serves the Object role in a finite clause; *dream* is an object in a nonfinite clause; *night* is an adverbial. Because only boundaries separating finite clauses are indicated, there is only one sentence-internal clause boundary, "@/" between *daughters* and *that*.

This kind of representation seeks to be (i) sufficiently expressive for stating grammatical generalisations in an economical and transparent fashion and (ii) sufficiently underspecific to make for a structurally resolvable grammatical representation. For example, the present way of functionally accounting for clauses enables the grammarian to express rules about the coordination of formally different but functionally similar entities. Regarding the resolvability requirement, certain kinds of structurally unresolvable distinctions are never introduced. For instance, the premodifier tag @>N only indicates that its head is a nominal in the right hand context.

### 2.3.3 A sample rule

Here is a realistic implication rule that partially defines the form of prepositional phrases:

```
            PREP =>
                        _ . @ Coord,
                        _ ..PrepComp,
PassVChain.. <Deferred> . _,
PostModiCl.. <Deferred> . _,
WH-Question.. <Deferred> . _;
```

A preposition is followed by a coordination or a preposition complement (here hidden in the constant *..PrepComp* that accepts e.g. noun phrases, nonfinite clauses and nominal clauses), or it (as a 'deferred' preposition) is preceded by a passive verb chain *PassVChain..* or a postmodifying clause *PostModiCl..* (the main verb in a postmodifying clause is furnished with the postmodifier tag *N<@*) or of a WH-question (i.e. in the same clause, there is a WH-word). If the tag *PREP* occurs in none of the specified contexts, the sentence reading containing it is discarded.

A comprehensive parsing grammar is under development. Currently it accounts for all major syntactic structures of English, but in a somewhat underspecific fashion. Though the accuracy of the grammar at the level of syntactic analysis can still be considerably improved, the syntactic grammar is already capable of resolving morphological ambiguities left pending by ENGCG.

# 3  An experiment with part-of-speech disambiguation

The system was tested against a 38,202-word test corpus consisting of previously unseen journalistic, scientific and manual texts.

The finite-state parser, the last module in the system, can in principle be "forced" to produce an unambiguous analysis for each input sentence, even for ungrammatical ones. In practice, the present implementation sometimes fails to give an analysis to heavily ambiguous inputs, regardless of their grammaticality.[6] Therefore two kinds of output were accepted for the evaluation: (i) the unambiguous analyses actually proposed by the finite-state parser, and (ii) the ENGCG analysis of those sentences for which the finite-state parser gave no analyses. From this nearly unambiguous combined output, the success of the hybrid was measured, by automatically comparing it with a benchmark version of the test corpus at the level of morphological (including part-of-speech) analysis (i.e. the syntax tags were ignored).

## 3.1  Creation of benchmark corpus

The benchmark corpus was created by first applying the preprocessor and morphological analyser to the test text. This morphologically analysed ambiguous text was then independently disambiguated by two experts whose task also was to detect any errors potentially produced by the previously applied components. They worked independently, consulting written documentation of the grammatical representation when necessary. Then these manually disambiguated versions were automatically compared. At this stage, slightly over 99% of all analyses were identical. When the differences were collectively examined, it was agreed that virtually all were due to inattention.[7] One of these two corpus versions was modified to represent the consensus, and this 'consensus corpus' was used as the benchmark in the evaluation.[8]

---

[6]During the intersection, the sentence automaton sometimes becomes prohibitively large.

|  | ambiguous words | readings | readings/word | errors | error rate |
|---|---|---|---|---|---|
| D0 (Morph. analysis) | 39.0% | 67,737 | 1.77 | 31 | 0.08% |
| D1 (D0 + ENGCG) | 6.2% | 40,450 | 1.06 | 124 | 0.32% |
| D2 (D1 + ENGCG heur.) | 3.2% | 38,949 | 1.02 | 226 | 0.59% |
| D3 (D2 + FS parser) | 0.6% | 38,342 | 1.00 | 281 | 0.74% |

Figure 1: Results from a tagging test on a 38,202-word corpus.

## 3.2 Results

The results are given in Figure 1.

Let us examine the results. ENGCG accuracy was close to normal, except that the heuristic constraints (tagger D2) performed somewhat poorer than usual.

The finite-state parser gave an analysis to about 80% of all words. Overall, 0.6% of all words remained ambiguous (due to the failure of the Finite State parser; c.f. Section 3). Parsing speed varied greatly (0.1-150 words/sec.) – refinement of the Finite State software is still underway.

The overall success of the system is very encouraging – 99.26% of all words retained the correct morphological analysis. Compared to the 95–97% accuracy of the best competing probabilistic part-of-speech taggers, this accuracy, achieved with an entirely rule-based description, suggests that part-of-speech disambiguation is a syntactic problem.

The misanalyses have not been studied in detail, but some general observations can be made:

- Many misanalyses made by the Finite State parser were due to ENGCG misanalyses (the "domino effect").

- The choice between adverbs and other categories was sometimes difficult. The distributions of adverbs and certain other categories overlaps; this may explain this error type. Lexeme-oriented constraints could be formulated for some of these cases.

---

[7]Only in the analysis of a few headings, different (meaning-level) interpretations arose, and even here it was agreed by both judges that this ambiguity was genuine.

[8]If this high consensus level appears surprising, see Voutilainen and Järvinen (this volume).

- Some ambiguities, e.g. noun–verb and participle–past tense, were problematic. This is probably due to the fact that while the parsing grammar always requires a regent for a dependent, it is much more permissive on dependentless regents. Clause boundaries, and hence the internal structure of clauses, could probably be determined more accurately if the heuristic part of the grammar also contained rules for preferring e.g. verbs with typical complements over verbs without complements.

# 4 Conclusion

Part-of-speech disambiguation has recently been tackled best with data-driven techniques. Linguistic techniques have done well at related levels (morphology, syntax) but not here. Is there something in parts of speech that makes them less accessible to the rule-based linguistic approach?

This paper outlines and evaluates a new part-of-speech tagger. It uses only linguistic distributional rules, yet reaches an accuracy clearly better than any competing system. This suggests that also parts of speech are a rule-governed distributional phenomenon.

The tagger has two rule components. One is a grammar specifically developed for resolution of part-of-speech ambiguities. Though much effort was given to its development, it leaves many ambiguities unresolved. These rules, superficially about parts of speech, actually express essentially syntactic generalisations, though indirectly and partially. The other rule component is a syntactic grammar. This syntactic grammar is able to resolve the pending part-of-speech ambiguities as a side effect.

In short: like morphology and syntax, parts of speech seem to be a rule-governed phenomenon. However, the best distributional account of parts of speech appears achievable by means of a syntactic grammar.[9]

# Acknowledgements

I would like to thank Timo Järvinen, Jussi Piitulainen, Pasi Tapanainen and two EACL referees for useful comments on an earlier version of this paper. The usual disclaimers hold.

---

[9]However, the parsing description would also benefit from a large corpus-based lexicon extension of compound nouns and other useful collocations for resolving some even syntactically genuine part-of-speech ambiguities. Collocations can be extracted from corpora using ENGCG-style corpus tools, e.g. *NPtool* (Voutilainen 1993).

# References

Hiyan Alshawi (ed.) 1992. *The Core Language Engine.* Cambridge, Mass.: The MIT Press.

Ezra Black, Fred Jelinek, John Lafferty, Robert Mercer and Salim Roukos 1992. Decision-tree models applied to the labeling of text with parts-of-speech. *Proceedings of the Workshop on Speech and natural Language.* Defence Advanced Research Projects Agency, U.S. Govt.

Eric Brill 1992. A simple rule-based part of speech tagger. *Proceedings of the Third Conference on Applied Natural Language Processing, ACL.*

Kenneth Church 1988. A Stochastic Parts Program and Noun Phrase Parser for Unrestricted Text. *Proceedings of the Second Conference on Applied Natural Language Processing, ACL.*

— 1992. Current Practice in Part of Speech Tagging and Suggestions for the Future. In Simmons (ed.), *Sbornik praci: In Honor of Henry Kučera.* Michigan Slavic Studies.

Douglass Cutting 1994. Porting a stochastic part-of-speech tagger to Swedish. In Eklund (ed.). 65-70.

Douglass Cutting, Julian Kupiec, Jan Pedersen and Penelope Sibun 1992. A Practical Part-of-Speech Tagger. *Proceedings of ANLP-92.*

Stephen DeRose 1988. Grammatical category disambiguation by statistical optimization. *Computational Linguistics.*

Robert Eklund (ed.) *Proceedings of '9:e Nordiska Datalingvistikdagarna', Stockholm 3-5 June 1993.* Department of Linguistics, Computational Linguistics, Stockholm University. Stockholm.

Martin Eineborg and Björn Gambäck 1994. Tagging experiment using neural networks. In Eklund (ed.). 71-81.

David Elworthy 1994. Does Baum-Welch re-estimation help taggers? In *Proceedings of the 4th Conference on Applied Natural Language Processing, ACL.* Stuttgart.

Elizabeth Eyes and Geoffrey Leech 1993. Syntactic Annotation: Linguistic Aspects of Grammatical Tagging and Skeleton Parsing. In Black *et al.* (eds.), *Statistically-Driven Computer Grammars of English: The IBM/Lancaster Approach.* Amsterdam: Rodopi.

Roger Garside 1987. The CLAWS word-tagging system. In Garside, Leech and Sampson (eds.), *The Computational Analysis of English.* London and New York: Longman.

Zelal Güngördü and Kemal Oflazer 1994. Parsing Turkish using the Lexical-Functional Grammar formalism. *Proceedings of COLING-94*, Vol. 1. Kyoto, Japan. 494-500.

Donald Hindle 1983. "User manual for Fidditch". Technical memorandum 7590-142, Naval Research Lab. USA.

— 1989. Acquiring disambiguation rules from text. *Proceedings of ACL-89*.

Karen Jensen, George Heidorn and Stephen Richardson (eds.) 1993. *Natural language processing: the PLNLP approach*. Kluver Academic Publishers: Boston.

Fred Karlsson, Atro Voutilainen, Juha Heikkilä and Arto Anttila (eds.) 1995. *Constraint Grammar. A Language-Independent System for Parsing Unrestricted Text*. Berlin and New York: Mouton de Gruyter.

Kimmo Koskenniemi 1983. *Two-level Morphology. A General Computational Model for Word-form Production and Generation*. Publications 11, Department of General Linguistics, University of Helsinki.

— 1990. Finite-state parsing and disambiguation. *Proceedings of the fourteenth International Conference on Computational Linguistics. COLING-90*. Helsinki, Finland.

Kimmo Koskenniemi, Pasi Tapanainen and Atro Voutilainen 1992. Compiling and using finite-state syntactic rules. In *Proceedings of the fifteenth International Conference on Computational Linguistics. COLING-92*. Vol. I, pp 156-162, Nantes, France.

Geoffrey Leech, Roger Garside and Michael Bryant 1994. CLAWS4: The tagging of the British National Corpus. In *Proceedings of COLING-94*. Kyoto, Japan.

Carl de Marcken 1990. Parsing the LOB Corpus. *Proceedings of the 28th Annual Meeting of the ACL*.

Mitchell Marcus, Beatrice Santorini and Mary Ann Marcinkiewicz 1993. Building a Large Annotated Corpus of English: The Penn Treebank. *Computational Linguistics*, Vol. 19, Number 2. 313-330.

Ian Marshall 1983. Choice of grammatical word-class without global syntactic analysis: tagging words in the LOB Corpus. *Computers in the Humanities*.

Michael McCord 1990. A System for Simpler Construction of Practical Natural Language Grammars. In R. Studer (ed.), *Natural Language and Logic. Lecture Notes in Artificial Intelligence 459*. Berlin: Springer Verlag.

Bernard Merialdo 1994. Tagging English text with a probabilistic model. *Computational Linguistics*, Vol. 20.

Geoffrey Sampson 1987. Probabilistic Models of Analysis. In Garside, Leech and Sampson (eds.).

Christer Samuelsson 1994. Morphological tagging based entirely on Bayesian inference. In Eklund (ed.). 225-237.

Helmut Schmid 1994. Part-of-speech tagging with neural networks. In *Proceedings of COLING-94*. Kyoto, Japan.

Daniel Sleator and Davy Temperley 1991. "Parsing English with a Link Grammar". CMU-CS-91-196. School of Computer Science, Carnegie Mellon University, Pittsburgh, PA 15213.

Tomek Strzalkowski 1992. TTP: a fast and robust parser for natural language. *Proceedings of the fifteenth International Conference on Computational Linguistics. COLING-92*. Nantes, France.

Pasi Tapanainen 1992. "Äärellisiin automaatteihin perustuva luonnollisen kielen jäsennin" (A finite state parser of natural language). Licentiate (pre-doctoral) thesis. Department of Computer Science, University of Helsinki.

Pasi Tapanainen and Timo Järvinen 1994. Syntactic analysis of natural language using linguistic rules and corpus-based patterns. *Proceedings of COLING-94*. Kyoto, Japan.

Pasi Tapanainen and Atro Voutilainen 1994. Tagging accurately – Don't guess if you know. *Proceedings of the 4th Conference on Applied Natural Language Processing, ACL*. Stuttgart.

Atro Voutilainen 1993. *NPtool*, a Detector of English Noun Phrases. In *Proceedings of the Workshop on Very Large Corpora*. Ohio State University, Ohio. 42-51.

— 1994. *Three studies of grammar-based surface parsing of unrestricted English text*. (Doctoral dissertation.). Publications 24, Department of General Linguistics, University of Helsinki.

Atro Voutilainen, Juha Heikkilä and Arto Anttila 1992. *Constraint Grammar of English. A Performance-Oriented Introduction*. Publications 21, Department of General Linguistics, University of Helsinki.

Atro Voutilainen and Pasi Tapanainen 1993. Ambiguity Resolution in a Reductionistic Parser. *Proceedings of the Sixth Conference of the European Chapter of the Association for Computational Linguistics*. Association for Computational Linguistics. Utrecht. 394-403.

Ralph Weischedel, Marie Meteer, Richard Schwartz, Lance Ramshaw and Jeff Palmuzzi 1993. Coping with ambiguity and unknown words through probabilistic models. *Computational Linguistics*, Vol. 19, Number 2.

# Appendix

Enclosed is a sample output of the system. Syntax tags have been retained; base forms and some tags have been removed for better readability. The syntactic tags used here are the following:

- **@>A**   premodifier of adjective, adverb or quantifier,
- **@>N**   noun premodifier,
- **@N<**   noun postmodifier,
- **@ADVL**   adverbial,
- **@ADVL/N<**   adverbial or noun postmodifier,
- **@OBJ**   object in a finite clause,
- **@IOBJ**   indirect object in a finite clause,
- **@SUBJ**   subject in a finite clause,
- **@obj**   object in a nonfinite clause,
- **@P<<**   preposition complement,
- **@nh**   nominal head,
- **@CC**   coordinating conjunction,
- **@CS**   subordinating conjunction,
- **@MV**   main verb in a finite clause,
- **@aux**   auxiliary in a nonfinite clause,
- **@mv**   main verb in a nonfinite clause,
- **ADVL@**   adverbial clause,
- **MC@**   finite main clause,
- **OBJ@**   clause as an object in a finite clause.

@@ On PREP @ADVL @
completion N NOM SG @P<< @
@comma @
check V IMP @MV MC@ @
the DET CENTRAL SG/PL @>N @
engine N NOM SG @>N @
oil N NOM SG @>N @
level N NOM SG @OBJ @/
@comma @
start V IMP @MV MC@ @
the DET CENTRAL SG/PL @>N @
engine N NOM SG @OBJ @/
then ADV ADVL @ADVL @
check V IMP @MV MC@ @
for PREP @ADVL @
oil N NOM SG @>N @

leaks N NOM PL @P<< @
@fullstop @@

@@ Screw V IMP @MV MC@ @
a DET CENTRAL SG @>N @
self-tapping PCP1 @>N @
screw N NOM SG @OBJ @
of PREP @N< @
appropriate A ABS @>N @
diameter N NOM SG @P<< @
into PREP @ADVL/N< @
this DET CENTRAL DEM SG @>N @
hole N NOM SG @P<< @/
@comma @
then ADV ADVL @ADVL @
lever V IMP @MV MC@ @
against PREP @ADVL @
the DET CENTRAL SG/PL @>N @
screw N NOM SG @P<< @
to INFMARK> @aux @
extract V INF @mv ADVL@ @
the DET CENTRAL SG/PL @>N @
plug N NOM SG @obj @
as CS @CS @
shown PCP2 @mv ADVL@ @
in PREP @ADVL @
FIG ABBR NOM SG @>N @
1.26 NUM CARD @P<< @
@fullstop @@

@@ This PRON DEM SG @nh @
done PCP2 @N< @
@comma @
push V IMP @MV MC@ @
the DET CENTRAL SG/PL @>N @
crankshaft N NOM SG @OBJ @
fully ADV @>A @
rearwards ADV @ADVL @/
@comma @
then ADV ADVL @ADVL @
slowly ADV @ADVL @
but CC @CC @
positively ADV @ADVL @
push V IMP @MV MC@ @

it PRON ACC SG3 @OBJ @
forwards ADV ADVL @ADVL @
to PREP @ADVL @
its PRON GEN SG3 @>N @
stop N NOM SG @P<< @
@fullstop @@

@@ Lightly ADV @ADVL @
moisten V IMP @MV MC@ @
the DET CENTRAL SG/PL @>N @
lips N NOM PL @OBJ @
of PREP @N< @
a DET CENTRAL SG @>N @
new A ABS @>N @
rear N NOM SG @>N @
oil N NOM SG @>N @
seal N NOM SG @P<< @
with PREP @ADVL/N< @
engine N NOM SG @>N @
oil N NOM SG @P<< @/
@comma @
then ADV ADVL @ADVL @
drive V IMP @MV MC@ @
it PRON ACC SG3 @OBJ @
squarely ADV @ADVL @
into PREP @ADVL @
position N NOM SG @P<< @/
until CS @CS @
it PRON NOM SG3 SUBJ @SUBJ @
rests V PRES SG3 @MV ADVL@ @
against PREP @ADVL @
its PRON GEN SG3 @>N @
abutment N NOM SG @P<< @
@comma @
preferably ADV @ADVL @
using PCP1 @mv ADVL@ @
the DET CENTRAL SG/PL @>N @
appropriate A ABS @>N @
service N NOM SG @>N @
tool N NOM SG @obj @
for PREP @ADVL/N< @
this DET CENTRAL DEM SG @>N @
operation N NOM SG @P<< @
@fullstop @@